\pdfoutput=1
\documentclass[11pt]{article}

\usepackage{emnlp2021}
\usepackage{times}
\usepackage{latexsym}

\usepackage[T1]{fontenc}
\usepackage[utf8]{inputenc}
\usepackage{amsmath, amssymb}
\usepackage[pdftex]{graphicx}
\usepackage{caption}
\usepackage{tikz}
\usepackage[capbesideposition=right,capbesidesep=quad]{floatrow}
\DeclareGraphicsExtensions{.pdf, .png}
\usepackage{microtype}

\title{OptAGAN: Entropy-based finetuning on text VAE-GAN}

\author{
  Paolo Tirotta\\
  Department of Statistics\\
  Alma Mater Studiorum\\
  University of Bologna \\
  \texttt{paolo.tirotta@gmail.com} \\
   \And
 Stefano Lodi \\
  Department of Computer Science\\
  Alma Mater Studiorum\\
  University of Bologna \\
  \texttt{stefano.lodi@unibo.it} \\
}

\begin{document}
\maketitle
\begin{abstract}
Transfer learning through large pre-trained models has changed the landscape of current applications in natural language processing (NLP). Recently Optimus, a variational autoencoder (VAE) which combines two pre-trained models, BERT and GPT-2, has been released, and its combination with generative adversial networks (GANs) has been shown to produce novel, yet very human-looking text.
The Optimus and GANs combination avoids the troublesome application of GANs to the discrete domain of text, and prevents the exposure bias of standard maximum likelihood methods. We combine the training of GANs in the latent space, with the finetuning of the decoder of Optimus for single word generation. This approach lets us model both the high-level features of the sentences, and the low-level word-by-word generation. We finetune using reinforcement learning (RL) by exploiting the structure of GPT-2 and by adding entropy-based intrinsically motivated rewards to balance between quality and diversity. We benchmark the results of the VAE-GAN model, and show the improvements brought by our RL finetuning on three widely used datasets for text generation, with results that greatly surpass the current state-of-the-art for the quality of the generated texts.
\end{abstract}

\section{Introduction}
\label{sec:introduction}
Unsupervised text generation finds its use on a plethora of real-world application, ranging from machine translation \cite{cit:mactrans}, to summarization \cite{cit:summ} and dialogue generation \cite{cit:diaggen}. A general approach to modelling text sequences is to autoregressively generate the next token given the previous ones, and the most successful and widespread technique is to train a model using maximum likelihood estimation (MLE). This approach, however, is not without fault. At training time the model learns to generate a token given the ground truth, while at inference time it takes as input its own generated sequence of words. This dissimilarity leads to the so-called exposure bias \cite{cit:expbias}, where the accumulation of errors during inference can produce poor outputs. Furthermore, the loss function of MLE is very strict. For each sequence, only the token accounted by the training sample is considered as correct \cite{cit:mleloss}, and the model learns precisely to mimic the given samples, often leading to quite dull and homogeneous outputs.

An alternative to MLE methods are generative adversial networks (GANs) \cite{cit:GAN}, where a generator learns to create outputs that can fool a discriminator into believing they are real. Thus, GANs do not have the strict loss function of MLE, and do not suffer from exposure bias, as they learn to sample during training. Nonetheless, the application of GANs to the text realm has been rather complicated. Due to the discreteness of text, the sampling of each token results in a non-differentiable function, which does not allow to backpropagate the loss of the discriminator. Countermeasures include the use of reinforcement learning (RL) \cite{cit:seqgan, cit:rankgan, cit:leakgan, cit:scratchgan}, the use of the Gumbel-Softmax relaxation \cite{cit:GSGAN, cit:relgan}, or to avoid the discrete space altogether and work with continuous embeddings using autoencoders \cite{cit:arae, cit:aegan, cit:latext}. However, methods which utilize RL  often rely on MLE pre-training, and usually do not improve over them \cite{cit:languagegans}.
Instead, for both the approaches using the Gumbel-Softmax distribution, and even more so for autoencoders, the discriminator considers a continuous representation of text, so it is not able to judge effectively the single word-by-word generation.

In the past few years, natural language processing (NLP) applications have found huge improvements with the introduction of the attention mechanism and the transformer architecture, with notable examples of BERT, GPT-2 and GPT-3 among others \cite{cit:transformer, cit:bert, cit:gpt2, cit:gpt3}. These kind of language models are large deep neural networks that are able to understand the depedencies between words thanks to attention and are trained over huge amounts of unannotated data. As such, pre-trained language models provide better language understanding over recurrent neural networks, can be very easily finetuned on a downstream task, including text generation, and reached state-of-the-art results in many areas. Recently Optimus, a text variational autoencoder (VAE), that is an autoencoder which maps sentences to a meaningful latent space, has been proposed \cite{cit:optimus}. It combines both BERT and GPT-2, as encoder and decoder respectively, and can be employed both as a generative model, and as a tool for language understanding tasks.

In this work, we aim to benchmark the results obtained from combining Optimus and GANs, similarly as indicated in the original paper. In doing so, we also investigate the GAN structure and compare the adaptive update strategy presented in \cite{cit:adapupdate} with the standard update strategy of GANs. Furthermore, we combine the training in the continuous space, with the finetuning of the decoder of Optimus in the discrete text space, in a similar fashion as done in ConCreteGAN \cite{cit:concretegan}. However, differently from most approaches which use RL, we do not use REINFORCE, but add an additional value head to GPT-2, which outputs the intermediate rewards \cite{cit:valuehead}. Moreover, we modify the reward function by considering the entropy of the model when generating tokens, and favour diversity in the output by adding an intrinsic reward.

Thus, our model OptAGAN \footnote{Opt(imus) A(ugmented) GAN - Implementation can be found at \href{https://github.com/Egojr/optagan}{https://github.com/Egojr/optagan}} is able to model both the higher level sentence structure, and has more control over single word generation, in a way that favours both quality and diversity for the generated sentences. We measure such criteria using standard automatic metrics: BLEU for quality, Backwards-BLEU for diversity, and Fréchet distance, on which we also present further analysis. We consider the image caption dataset COCO, the Stanford Natural Language Inference (SNLI) dataset, and the EMNLP News 2017 dataset for unconditional text generation, and also provide results for the conditional review dataset YELP.

Results show that the base VAE-GAN model already improves over other GAN methods, especially with regards to quality. OptAGAN, further improves over these results, and manages to handle the quality-diversity tradeoff very well. Moreover, we show a further experiment that helps understanding the strenghts and weaknesses of our finetuning approach.

\section{Background}\label{sec:background}
In this section we introduce the mathematical notation and briefly describe the main theoretical tools which are used in OptAGAN. We also present an overview of the other methods of text generation.

\subsection{Variational autoencoders}
VAEs are generative models formed by two independent models, an encoder $q_{\phi}$ and a decoder $p_{\theta}$. The encoder is tasked with mapping the input $\boldsymbol{x}$ to a latent space $\boldsymbol{z}$ that allows for interpolation. The decoder maps from $\boldsymbol{z} \rightarrow \Tilde{\boldsymbol{x}}$, providing an approximation of the original input. Thanks to the introduction of a local variation from sampling the encoder output, it is possible to induce a smooth latent representation of the inputs, which differs from the rigid space of autoencoders. 
\paragraph{Optimus} Optimus combines the autoregressive nature of the GPT-2 text generation with the latent produced by the encoder, such that text generation is done as:
\begin{equation}
    p_{\boldsymbol{\theta}}(\boldsymbol{x}|\boldsymbol{z})=\prod_{t=1}^{n}p_{\boldsymbol{\theta}}(x_{t}|x_{1},...,x_{t-1},\boldsymbol{z}),
\end{equation}
where the probability of each token is estimated conditionally on the latent embedding and the previous tokens. The latent vector $\boldsymbol{z}$, which comes from the output of the BERT encoder, controls the high-level characteristics of the sentence, such as length, tense, style and topic, and allows for the guided generation of text.

\subsection{Generative adversarial networks}
GANs are also generative models formed by two models: a generator and a discriminator. Differently from VAEs, the generator $G$ samples from a random variable to produce output that can fool the discriminator $D$ into believing they are real, while the discriminator is constantly learning to distinguish between the real and generated data. The objective of the two models can formulated as:
\begin{equation}
\begin{split}
     \min_{G}\max_{D}& V(D,G)= \mathbb{E}_{\boldsymbol{x} \sim p_{D}}[\log (D(\boldsymbol{x}))] \\
     &+ \mathbb{E}_{\boldsymbol{\epsilon} \sim p_{\epsilon}}[ \log (1-D(G(\boldsymbol{\epsilon}))],
\end{split}
\end{equation}

where $p_{D}$ and $p_{\epsilon}$ are the distribution of the data and of the input noise of the generator, respectively. When combining GANs with autoregressive text generation, the operation of sampling the next tokens is non-differentiable, so the application of GANs relies on either the use of policy gradient algorithms, or the use of continuous approximations, such as the combination of GANs with autoencoders or VAEs.

\subsection{Reinforcement learning}
Approaches that use policy gradient algorithms to allow the training of GANs consider the generator as the policy $\pi_{\theta}$ to train, the sampling of the token as an action $A$ from a state $S$, and the output of the discriminator as the reward $R$. 
For each sentence generated by a model a reward is calculated, however, as the discriminator only calculates the reward over finished sentences, the intermediate rewards, usually referring to the word-by-word generation, are obtained through the REINFORCE algorithm and Monte-Carlo rollout. Optimization of the parameters is performed through gradient ascent:
\begin{gather}
    \boldsymbol{\theta}_{t+1} = \boldsymbol{\theta}_{t} + \alpha \nabla_{\boldsymbol{\theta}} J(\boldsymbol{\theta}_{t})  \\
    \nabla_{\boldsymbol{\theta}} J(\boldsymbol{\theta}_{t}) \propto \mathbb{E}_{\pi} \left[ G_{t} \nabla_{\boldsymbol{\theta}} \ln \pi(A_{t}|S_{t},\boldsymbol{\theta}) \right],  \label{eq:reinforce}
\end{gather}
where the gradient of the REINFORCE objective, $\nabla_{\boldsymbol{\theta}} J(\boldsymbol{\theta}_{t})$ is proportional to the discounted returns $G_{t}=\sum_{j=0}^{t}\gamma^{j}R_{j}$, so that higher return actions are favored, and is inversely proportional to the probability of being selected, so that higher probability actions are not at an advantage compared to low probability ones. Equation \ref{eq:reinforce} also provides a value that can be sampled at each time step and only depends on the policy $\pi$.

\subsection{Related work}
Many works have dealt with the training of GANs in the discrete realm, starting with SeqGAN \cite{cit:seqgan}, LeakGAN \cite{cit:leakgan} and RankGAN \cite{cit:rankgan}, where all of them share a similar structure, mostly differing in the form of the discriminator, and require MLE pre-training followed by adversarial training with REINFORCE. ScratchGAN \cite{cit:scratchgan} is the first model to show that MLE pre-training can be avoided by carefully combining existing techniques. Other works that train GANs using continuous relaxations include ARAE \cite{cit:arae} and LATEXT-GAN \cite{cit:latext}, which use autoencoders to learn a continuous latent representation. Models based on the Gumbel-Softmax distribution are RelGAN \cite{cit:relgan} and GSGAN \cite{cit:GSGAN}. On the comparison of these methods and the evaluation metrics, \cite{cit:languagegans, cit:metrics} have shown the inadequacy of current GANs when compared to MLE and the need for metrics that can better measure the quality and diversity of the models. 
\\
On the topic of exploration of text GAN models and RL are ColdGANs \cite{cit:coldgans}, which delve deeper into the effects of temperature for the text generation. An approach similar to ours, which involves the use of large pre-trained models and RL is TextGAIL \cite{cit:textgail}, where both GPT-2 as the generator, and RoBERTa as the discriminator are used for the task of text generation.
\\
Regarding text VAEs, Optimus \cite{cit:optimus} is the first large pre-trained model of such kind, whereas previously researchers had tried developping VAEs using either recurrent neural networks \cite{cit:oldvae} or semi-amortized inference \cite{cit:semiamortized}.

\section{OptAGAN}

\begin{figure*}[h!]
\centering
\resizebox{\columnwidth}{!}{
\tikzset{every picture/.style={line width=0.75pt}} %set default line width to 0.75pt        

\begin{tikzpicture}[x=0.75pt,y=0.75pt,yscale=-1,xscale=1]
%uncomment if require: \path (0,253); %set diagram left start at 0, and has height of 253

%Shape: Rectangle [id:dp8564035009487291] 
\draw   (20.7,40.57) -- (49.7,40.57) -- (49.7,70.55) -- (20.7,70.55) -- cycle ;
%Shape: Rectangle [id:dp5444178157371958] 
\draw   (139.7,40.63) -- (168.7,40.63) -- (168.7,70.55) -- (139.7,70.55) -- cycle ;
%Straight Lines [id:da4253569105887146] 
\draw    (49.67,55.33) -- (137.73,55.66) ;
\draw [shift={(139.73,55.67)}, rotate = 180.21] [color={rgb, 255:red, 0; green, 0; blue, 0 }  ][line width=0.75]    (10.93,-3.29) .. controls (6.95,-1.4) and (3.31,-0.3) .. (0,0) .. controls (3.31,0.3) and (6.95,1.4) .. (10.93,3.29)   ;
%Straight Lines [id:da001247253957682215] 
\draw    (169.47,56.33) -- (257.53,56.66) ;
\draw [shift={(259.53,56.67)}, rotate = 180.21] [color={rgb, 255:red, 0; green, 0; blue, 0 }  ][line width=0.75]    (10.93,-3.29) .. controls (6.95,-1.4) and (3.31,-0.3) .. (0,0) .. controls (3.31,0.3) and (6.95,1.4) .. (10.93,3.29)   ;
%Shape: Rectangle [id:dp4419766093521831] 
\draw   (260.7,40.73) -- (289.7,40.73) -- (289.7,70.55) -- (260.7,70.55) -- cycle ;
%Shape: Rectangle [id:dp11539103006616425] 
\draw  [color={rgb, 255:red, 208; green, 2; blue, 27 }  ,draw opacity=1 ] (20.7,130.55) -- (49.67,130.55) -- (49.67,160.55) -- (20.7,160.55) -- cycle ;
%Shape: Rectangle [id:dp5105286115205083] 
\draw  [color={rgb, 255:red, 208; green, 2; blue, 27 }  ,draw opacity=1 ] (140.7,130.63) -- (169.67,130.63) -- (169.67,160.55) -- (140.7,160.55) -- cycle ;
%Straight Lines [id:da28223864758696027] 
\draw [color={rgb, 255:red, 208; green, 2; blue, 27 }  ,draw opacity=1 ]   (49.67,146.33) -- (137.73,146.66) ;
\draw [shift={(139.73,146.67)}, rotate = 180.21] [color={rgb, 255:red, 208; green, 2; blue, 27 }  ,draw opacity=1 ][line width=0.75]    (10.93,-3.29) .. controls (6.95,-1.4) and (3.31,-0.3) .. (0,0) .. controls (3.31,0.3) and (6.95,1.4) .. (10.93,3.29)   ;
%Straight Lines [id:da44259360340870235] 
\draw [color={rgb, 255:red, 208; green, 2; blue, 27 }  ,draw opacity=1 ]   (170.47,146.33) -- (258.53,146.66) ;
\draw [shift={(260.53,146.67)}, rotate = 180.21] [color={rgb, 255:red, 208; green, 2; blue, 27 }  ,draw opacity=1 ][line width=0.75]    (10.93,-3.29) .. controls (6.95,-1.4) and (3.31,-0.3) .. (0,0) .. controls (3.31,0.3) and (6.95,1.4) .. (10.93,3.29)   ;
%Shape: Rectangle [id:dp5483715299698102] 
\draw  [color={rgb, 255:red, 208; green, 2; blue, 27 }  ,draw opacity=1 ] (261.7,130.63) -- (290.67,130.63) -- (290.67,160.55) -- (261.7,160.55) -- cycle ;
%Shape: Rectangle [id:dp1684155746076662] 
\draw  [color={rgb, 255:red, 0; green, 0; blue, 0 }  ,draw opacity=1 ][dash pattern={on 0.84pt off 2.51pt}] (125.93,27.57) -- (184.27,27.57) -- (184.27,177.64) -- (125.93,177.64) -- cycle ;
%Straight Lines [id:da5270475579000684] 
\draw [color={rgb, 255:red, 74; green, 144; blue, 226 }  ,draw opacity=1 ]   (290.47,146.53) -- (378.53,146.86) ;
\draw [shift={(380.53,146.87)}, rotate = 180.21] [color={rgb, 255:red, 74; green, 144; blue, 226 }  ,draw opacity=1 ][line width=0.75]    (10.93,-3.29) .. controls (6.95,-1.4) and (3.31,-0.3) .. (0,0) .. controls (3.31,0.3) and (6.95,1.4) .. (10.93,3.29)   ;
%Shape: Rectangle [id:dp21384448659039434] 
\draw  [color={rgb, 255:red, 74; green, 144; blue, 226 }  ,draw opacity=1 ] (379.7,130.63) -- (408.67,130.63) -- (408.67,160.55) -- (379.7,160.55) -- cycle ;
%Straight Lines [id:da6967534014875817] 
\draw [color={rgb, 255:red, 0; green, 0; blue, 0 }  ,draw opacity=1 ]   (154.2,178.44) -- (188.31,207.31) ;
\draw [shift={(189.84,208.6)}, rotate = 220.24] [color={rgb, 255:red, 0; green, 0; blue, 0 }  ,draw opacity=1 ][line width=0.75]    (10.93,-3.29) .. controls (6.95,-1.4) and (3.31,-0.3) .. (0,0) .. controls (3.31,0.3) and (6.95,1.4) .. (10.93,3.29)   ;
%Straight Lines [id:da05179181950168976] 
\draw [color={rgb, 255:red, 0; green, 0; blue, 0 }  ,draw opacity=1 ]   (154.2,178.44) -- (120.17,206.92) ;
\draw [shift={(118.64,208.2)}, rotate = 320.07] [color={rgb, 255:red, 0; green, 0; blue, 0 }  ,draw opacity=1 ][line width=0.75]    (10.93,-3.29) .. controls (6.95,-1.4) and (3.31,-0.3) .. (0,0) .. controls (3.31,0.3) and (6.95,1.4) .. (10.93,3.29)   ;
%Curve Lines [id:da8184998579172483] 
\draw [color={rgb, 255:red, 74; green, 144; blue, 226 }  ,draw opacity=1 ]   (408.87,146.9) .. controls (449.53,149.57) and (458.87,100.57) .. (408.88,100.4) ;
%Curve Lines [id:da5389206916064531] 
\draw [color={rgb, 255:red, 74; green, 144; blue, 226 }  ,draw opacity=1 ]   (276.88,100.4) .. controls (336.08,100) and (369.28,100.4) .. (408.88,100.4) ;
%Curve Lines [id:da06345196626981342] 
\draw [color={rgb, 255:red, 74; green, 144; blue, 226 }  ,draw opacity=1 ]   (276.88,100.4) .. controls (215.26,100.4) and (219.04,104.58) .. (216.3,128.89) ;
\draw [shift={(216.08,130.8)}, rotate = 277.02] [color={rgb, 255:red, 74; green, 144; blue, 226 }  ,draw opacity=1 ][line width=0.75]    (10.93,-3.29) .. controls (6.95,-1.4) and (3.31,-0.3) .. (0,0) .. controls (3.31,0.3) and (6.95,1.4) .. (10.93,3.29)   ;
%Shape: Rectangle [id:dp6740468427934228] 
\draw  [color={rgb, 255:red, 74; green, 144; blue, 226 }  ,draw opacity=1 ][dash pattern={on 0.84pt off 2.51pt}] (188.44,88.8) -- (459.87,88.8) -- (459.87,178.4) -- (188.44,178.4) -- cycle ;

% Text Node
\draw (269,49.4) node [anchor=north west][inner sep=0.75pt]    {$\tilde{x}$};
% Text Node
\draw (29.5,52.3) node [anchor=north west][inner sep=0.75pt]    {$x$};
% Text Node
\draw (149.4,52.5) node [anchor=north west][inner sep=0.75pt]    {$z$};
% Text Node
\draw (29.53,141.9) node [anchor=north west][inner sep=0.75pt]    {$\epsilon $};
% Text Node
\draw (149.93,139.83) node [anchor=north west][inner sep=0.75pt]    {$\hat{z}$};
% Text Node
\draw (270.2,139.23) node [anchor=north west][inner sep=0.75pt]    {$\hat{x}$};
% Text Node
\draw (386.93,139.36) node [anchor=north west][inner sep=0.75pt]    {$R$};
% Text Node
\draw (61,36.72) node [anchor=north west][inner sep=0.75pt]   [align=left] {Encoder};
% Text Node
\draw (187.53,37.92) node [anchor=north west][inner sep=0.75pt]   [align=left] {Decoder};
% Text Node
\draw (54.8,127.12) node [anchor=north west][inner sep=0.75pt]   [align=left] {Generator};
% Text Node
\draw (188.2,127.92) node [anchor=north west][inner sep=0.75pt]   [align=left] {Decoder};
% Text Node
\draw (313.2,128.32) node [anchor=north west][inner sep=0.75pt]   [align=left] {BLEU};
% Text Node
\draw (112.8,9.25) node [anchor=north west][inner sep=0.75pt]   [align=left] {Discriminator};
% Text Node
\draw (103.4,211.52) node [anchor=north west][inner sep=0.75pt]   [align=left] {Real};
% Text Node
\draw (174.6,211.52) node [anchor=north west][inner sep=0.75pt]   [align=left] {Fake};
% Text Node
\draw (482.4,33.16) node [anchor=north west][inner sep=0.75pt]    {$x$};
% Text Node
\draw (483,122.76) node [anchor=north west][inner sep=0.75pt]    {$z$};
% Text Node
\draw (481,60.16) node [anchor=north west][inner sep=0.75pt]    {$\tilde{x}$};
% Text Node
\draw (484.8,183.8) node [anchor=north west][inner sep=0.75pt]    {$\epsilon $};
% Text Node
\draw (483.4,149.36) node [anchor=north west][inner sep=0.75pt]    {$\hat{z}$};
% Text Node
\draw (482.2,89.76) node [anchor=north west][inner sep=0.75pt]    {$\hat{x}$};
% Text Node
\draw (481.8,210.16) node [anchor=north west][inner sep=0.75pt]    {$R$};
% Text Node
\draw (495.44,31.2) node [anchor=north west][inner sep=0.75pt]   [align=left] {= Original text};
% Text Node
\draw (494.84,61) node [anchor=north west][inner sep=0.75pt]   [align=left] {= Reconstructed text};
% Text Node
\draw (495.44,90.4) node [anchor=north west][inner sep=0.75pt]   [align=left] {= Generated text};
% Text Node
\draw (495.04,120.2) node [anchor=north west][inner sep=0.75pt]   [align=left] {= Latent variable};
% Text Node
\draw (495.24,150.6) node [anchor=north west][inner sep=0.75pt]   [align=left] {= Generated latent variable};
% Text Node
\draw (496.44,180.72) node [anchor=north west][inner sep=0.75pt]   [align=left] {= Noise};
% Text Node
\draw (497.44,209.32) node [anchor=north west][inner sep=0.75pt]   [align=left] {= Rewards};

\end{tikzpicture}
}
\caption{Structure of our proposed model OptAGAN. In red is the main process to generate new text, while in black we show the other models taking part in the training of OptAGAN. Finally, in blue is the last bit of the RL finetuning process.} \label{fig:optagan}
\end{figure*}
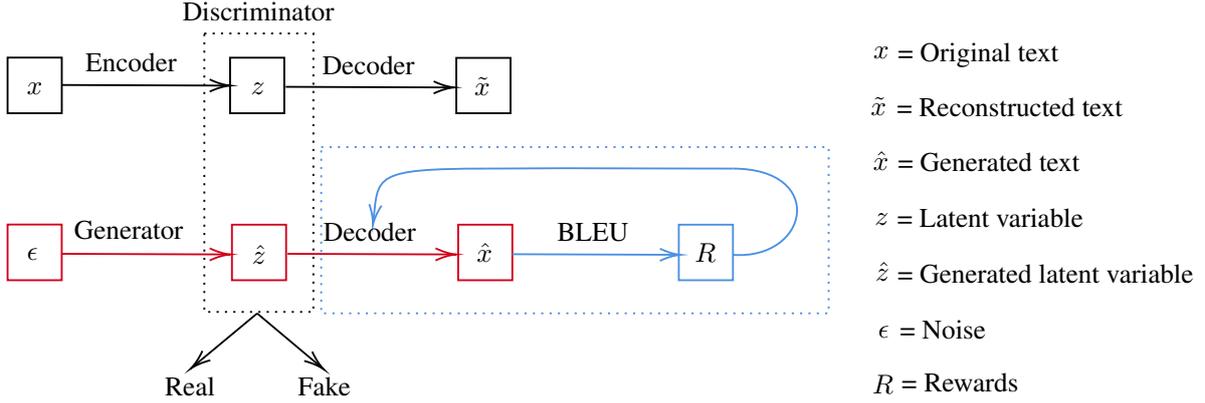

The architecture that we present in this work is composed of three main processes, as can be seen from Figure \ref{fig:optagan}. Each process is independent of the others, so each part is trained sequentially.
\begin{itemize}
    \item In order to fully utilize the strengths of Optimus, we finetune both the encoder and the decoder on the target dataset. The end results are a more separated and distinct latent space for each sentence, and a decoder which better reconstructs the original sentences.
    \item Next, we train the GAN model composed of the generator and the discriminator. In the case of conditional generation, we also add a classifier, whose loss is then passed to the generator. Both the generator and the discriminator only consider the continuous latent embeddings, so they are much lighter and faster to train compared to other text GANs.
    \item Finally, we finetune the decoder on discrete text using a value head, which estimates the reward of each single token in a sentence for the generated sequences. The estimated rewards are also augmented by considering the model entropy of each generated token. The gradient is then passed to the decoder through simple policy gradient. 
\end{itemize}
The structure of both the generator and the discriminator is a simple feed-forward neural network. Current literature does not give clear answers about which loss function specification for GANs is best for continuous data such as text latent embeddings. Our experiments show that Wasserstein GANs with gradient penalty (WGAN-GP) perform the best over sliced Wasserstein distances, which produce very homogeneous outputs. Thus, the loss function to optimize for the generator and discriminator is:
\begin{equation}
\begin{split}
    \min_{G} & \max_{D} \mathbb{E}_{\boldsymbol{x} \sim p_{D}}[D(\boldsymbol{x})] - \mathbb{E}_{\boldsymbol{\epsilon} \sim p(\epsilon)}[D(G(\boldsymbol{\epsilon}))] \\
    &+ \lambda \mathbb{E}_{\hat{\boldsymbol{x}} \sim p(\hat{x})} \left[ \|(\triangledown_{\hat{\boldsymbol{x}}}D(\hat{\boldsymbol{x}})\|_{2}-1)^{2} \right].
\end{split}
\end{equation}

\subsection{Update strategy}
When training GANs, the common update strategy is to have $k$ update steps, usually in the range $[5,10]$, for the discriminator for each step of the generator, as it has been shown to give stable training for GANs. We indeed also use this update method, but we also experiment with an adaptive update strategy proposed in \cite{cit:adapupdate}, where the choice to update the discriminator or the generator is given by a comparison of the loss change ratio of the two networks.
\begin{align}
    & r_{G} = \frac{|(L^{c}_{G}-L^{p}_{G})|}{L^{p}_{G}+c},
    & r_{D} = \frac{|(L^{c}_{D}-L^{p}_{D})|}{L^{p}_{D}+c},
\end{align}
where the relative change between the current loss $L^{c}$ and previous loss $L^{p}$ for both networks is used in determining which one gets updated. We also add an arbitrarily small constant $c$ in case the losses are too close to 0.
A weight $\lambda$ can be also introduced, so that if $r_{D}>\lambda r_{G}$ the discriminator is updated, and viceversa. Contrarily to the original paper, which suggests a value $\lambda \ge 1$, we notice that at the beginning of training there is a stark imbalance in the number of updates between the networks resulting in slower convergence. To better balance the training, we end up using a value of $\lambda < 1$, which converges to $1$ with each passing epoch of training.

\subsection{Value head}
Due to the high computational costs of implementing a text discriminator with a vocabulary of size equal to GPT-2, we rely on an external value head, whose scalar output for each token corresponds to the intermediate reward. Let the hidden states of the decoder $\rho$ and the value head $\nu$, the rewards calculated from an external metric $R^{ex}$ and each state $S_{t}$ are:
\begin{equation}
\begin{split}
    R_{t} = \nu(S_{t}|\rho), \quad & \text{with } \text{$\nu(S_{t}|\rho)$ minimizing} \\
    & \left|\sum_{t=0}^{T}(\nu(S_{t}|\rho))-R^{ex}\right| \label{eq:mae}.
\end{split}
\end{equation}
    
The external head takes as input the frozen hidden states of GPT-2. Freezing the hidden parameters is necessary because we are not modelling the parameters of the VAE model during RL. The loss is then calculated according to a MAE objective and passed back to the value head. This process estimates how much each token contributes to the reward, and comparatively to a text discriminator is faster to train and showed better results.

\subsection{Entropy-based rewards}
Our external rewards are calculated based on the quality metric BLEU, which we briefly describe in Section \ref{sec:metrics}. Under many considered reward specifications, which included the addition of diversity metrics, maximum entropy RL, changes of temperature, or a combination of these, the increase in quality is counterbalanced with a drop in the diversity of the generated sentences. One approach that managed to balance the quality-diversity trade-off was the addition of an intrinsically motivated penalty based on the confidence of the model when generating the token, calculated by the entropy. If again we consider the hidden states $\rho$, the last layer calculating the logits as our policy $\pi$ with parameters $\boldsymbol{\theta}$ and the entropy as $H(\pi(\cdot|\rho,S_{t}))$, we calculate the intrinsic rewards and the performance objective as:
\begin{align}
     R^{in}_{t} = \log( &\text{clamp($H(\pi(\cdot|\rho,S_{t})),0.2,1$)}) \label{eq:intrew} 
     \\[2ex]
\begin{split}
    \nabla_{\boldsymbol{\theta}} J(\boldsymbol{\theta}_{t}) = & \mathbb{E}_{\pi} \biggl[ \biggl( \sum_{k=0}^{t}(\gamma^{k} R_{k})+R^{in}_{t}\biggr) \\
    & \cdot \nabla_{\boldsymbol{\theta}} \ln \pi(A_{t}|\rho,S_{t},\boldsymbol{\theta}) \biggr] \label{eq:gradient}  
\end{split}
\end{align}
This specification favours high-reward actions with high entropy, while low-entropy actions have to have a high enough reward to be able to keep their high probability, resulting in a more diverse generation. As a rule of thumb, we found out that penalties should be lower than the maximum overall reward.

\section{Experimental settings}
In this section, we introduce the automatic metrics and the datasets used for evaluation. For comparison we consider a MLE model, SeqGAN, RankGAN, as implemented by the benchmarking platform Texygen \cite{cit:texygen}, and ScratchGAN. We also provide further details on our RL finetuning and present issues with the current evaluation metrics.

\subsection{Evaluation metrics}\label{sec:metrics}
BLEU is a metric that measures the overlapping n-grams between a hypothesis text and all the reference texts. The final score is calculated as the average of the scores over all hypothesis sentences. Studies have shown \cite{cit:metrics, cit:languagegans} that BLEU can only detect small syntax problems, resulting in poor correlation with human evaluations, however it still remains the standard when evaluating the quality of generated texts. To measure diversity we utilize Backwards-BLEU (BBLEU) where the generated texts are the reference and the test set becomes the hypothesis, giving a measure of how much the test set is represented.
Additionally, we consider the Fréchet Distance with the InferSent embedding model (FID). It has been shown that FID responds better than BLEU at identifying mode collapse and changes in words usage. However, we show that it can be biased due to its distributional assumptions, mainly for differences in sentence length distribution. Nonetheless, it can be useful in identifying very homogeneous outputs, especially in conjunction with BLEU and BBLEU scores.

\subsection{Datasets}
We consider three of most widely used datasets for unconditional text generation: image COCO \cite{cit:COCO}, Stanford Natural Language Inference (SNLI) \cite{cit:SNLI} and the EMNLP News 2017 dataset\footnote{http://www.statmt.org/wmt17/}. Moreover, we consider the YELP review dataset for conditional text generation \cite{cit:yelp}.
\begin{table}[h]
\centering
\resizebox{\textwidth}{!}{\begin{tabular}{c|c|c|c|c}
                         & COCO   & SNLI     & EMNLP   & YELP   \\
\hline\hline
Conditional              &  x & x   & x  & \checkmark  \\
\hline
Average sentence length  &  11.3  &  9.7     & 28.8    & 96.4  \\
\hline
Size of train set        &  10k   &  100k    & 270k    & 100k  \\
\hline
Size of dev set          &  10k   &  10k     & 10k     & 10k  \\
\hline
Size of test set         &  10k   &  10k     & 10k     & 10k  \\
\hline
\end{tabular}}
\caption[Table Datasets]{Average length of the train set and number of sentences for each of the datasets used for evaluation.}
\label{tab:Table data}
\end{table}
Each of the datasets presents different challenges when training: COCO and SNLI are a small and medium-sized dataset with short sentences, respectively. EMNLP is a large dataset with longer sentences. Lastly, the YELP dataset is a conditional, medium-sized dataset with very long sentences. The preprocessing on the datasets is minimal and limited to the YELP dataset.

\begin{table*}[h!]
\centering
\resizebox{\textwidth}{!}{\begin{tabular}{c|c|c|c|c|c|c|c}
 \hline
 Metrics & MLE & SeqGAN & RankGAN & ScratchGAN & Standard  & Adaptive & OptAGAN \\
 \hline
 BLEU-2 $\uparrow$  & 0.829  & 0.796  &  0.764 &  0.835 & 0.825 & 0.816 & \textbf{0.860} \\
 BLEU-3 $\uparrow$  & 0.548  & 0.471 &  0.399 &  0.556 & 0.554 & 0.544 & \textbf{0.605} \\
 BLEU-4 $\uparrow$  & 0.304  & 0.228 &  0.159 &  0.313 & 0.285 & 0.284 & \textbf{0.356} \\
 \hline \hline
 BBLEU-2 $\uparrow$ & 0.840  & 0.762  &  0.728 &  0.824 & 0.765 & 0.805 & \textbf{0.841}\\
 BBLEU-3 $\uparrow$ & 0.563  & 0.450  &  0.383 &  0.545 & 0.488 & 0.526 & \textbf{0.586} \\
 BBLEU-4 $\uparrow$ & 0.317  & 0.221  &  0.163 &  0.303  & 0.261 & 0.278 & \textbf{0.350} \\
  \hline \hline
 FID  $\downarrow$ & 0.926 & 1.934 & 3.509 & \textbf{0.466} & 1.153 & 0.784 & 0.674 
\end{tabular}}
\caption{EMNLP results of the automatic metrics for OptAGAN, the two base VAE-GAN models with standard and adaptive updates and the other models implemented for comparison.} 
\label{tab:Table models EMNLP}
\end{table*} 

\begin{table*}[h!]
\centering
\resizebox{\textwidth}{!}{\begin{tabular}{c|c|c|c|c|c|c|c}
 \hline
 Metrics & MLE & SeqGAN & RankGAN & ScratchGAN & Standard  & Adaptive & OptAGAN \\
 \hline
 BLEU-2 $\uparrow$ & 0.841  & 0.838  &  0.784 &  0.795 & 0.867 & 0.888 & \textbf{0.889} \\
 BLEU-3 $\uparrow$ & 0.635  & 0.599  &  0.514 &  0.564 & 0.693 & 0.726 & \textbf{0.727} \\
 BLEU-4 $\uparrow$ & 0.428  & 0.380  &  0.309 &  0.363 & 0.484 & 0.524 & \textbf{0.525} \\
 \hline \hline
 BBLEU-2 $\uparrow$ & \textbf{0.843}  & 0.768  &  0.771 &  0.800 & 0.786 & 0.764 & 0.764 \\
 BBLEU-3 $\uparrow$ & \textbf{0.639}  & 0.546  &  0.523 &  0.564 & 0.570 & 0.547 & 0.548 \\
 BBLEU-4 $\uparrow$ & \textbf{0.433}  & 0.347  &  0.347 &  0.362 & 0.374 & 0.361 & 0.362 \\
  \hline \hline
 FID $\downarrow$  & \textbf{0.376}  & 0.919  &  1.486 &  0.539 & 1.424 & 1.764 & 1.765
\end{tabular}}
\caption{SNLI results of the automatic metrics for OptAGAN, the two base VAE-GAN models with standard and adaptive updates and the other models implemented for comparison.}
\label{tab:Table models SNLI}
\end{table*} 

\begin{table*}[h!]
\centering
\resizebox{\textwidth}{!}{\begin{tabular}{c|c|c|c|c|c|c|c}
 \hline
 Metrics & MLE & SeqGAN & RankGAN & ScratchGAN & Standard  & Adaptive & OptAGAN \\
 \hline
 BLEU-2 $\uparrow$ & 0.854  & 0.866  &  0.834 &  0.862 & 0.917 & 0.919 & \textbf{0.920} \\
 BLEU-3 $\uparrow$ & 0.664  & 0.667  &  0.641 &  0.678 & 0.792 & 0.792 & \textbf{0.794} \\
 BLEU-4 $\uparrow$ &  0.459  & 0.452  &  0.423 &  0.478 & \textbf{0.627} &  0.617 & 0.620 \\
 \hline \hline
 BBLEU-2 $\uparrow$  & \textbf{0.844}  & 0.813  &  0.775 &  0.791 & 0.755 &  0.775 & 0.778 \\
 BBLEU-3 $\uparrow$  & \textbf{0.656}  & 0.616  &  0.553 &  0.573 & 0.550 &  0.579 & 0.584 \\
 BBLEU-4 $\uparrow$  & \textbf{0.455}  & 0.415  &  0.344 &  0.377 & 0.368 &  0.396 & 0.399 \\
  \hline \hline
 FID $\downarrow$  & \textbf{0.588}  & 0.756  &  2.347 &  1.001 & 2.297 &  1.908 & 1.903
\end{tabular}}
\caption{COCO results of the automatic metrics for OptAGAN, the two base VAE-GAN models with standard and adaptive updates and the other models implemented for comparison.}
\label{tab:Table models COCO}
\end{table*}

\subsection{Experimental results}

Tables \ref{tab:Table models EMNLP}, \ref{tab:Table models SNLI}, \ref{tab:Table models COCO} show the results for the quality and diversity of the models. The changes between standard and adaptive updates mostly favour the adaptive one, with larger gains in diversity, with the exception of the SNLI dataset. Additional considerations about the two updates can be found in Appendix \ref{app:adap}. Therefore we only apply the RL finetuning on the adaptively trained model to obtain the OptAGAN results. Our approach shows improvements under all metrics, albeit small for the COCO and SNLI datasets. In comparison with the other GAN models, OptAGAN boasts the highest quality, and average, or higher diversity. In comparison with the MLE model, the quality-diversity trade-off favours our model for quality, and the MLE approach for diversity. Notably, for the EMNLP dataset, the curiosity-driven finetuning allows OptAGAN to surpass all models for both BLEU and BBLEU.
We believe that the difference in the magnitude of change between the EMNLP task and the COCO and SNLI ones is due the starting quality of the model. In fact, as the RL finetuning slightly prioritizes quality, so increases in BLEU score, over diversity, the actual changes on the word-by-word generation are very few for those two datasets.
Compared to the other methods, ours is also better able to reproduce longer sequences of words, as the growing differences between 2,3 and 4 n-grams metrics show. Regarding the FID scores, they mostly show the same behaviour as BBLEU. However, we show in section \ref{sec:fid analysis} that the FID is biased for the sentence length distribution, that, in contrast with other methods such as ScratchGAN, we do not model.
From the computational cost point of view, the full training of our model took at most 24 hours using a single Tesla V100 GPU. Additional details about the training can be found in Appendix \ref{app:training}. 

\subsection{Conditional generation}
For the conditional generation task we follow the same procedure as the unconditional one, with the only exception of the addition of a classifier network to better model the GAN generation depending on the label. The results of table \ref{tab:Table YELP results} are very similar to the ones of the COCO and SNLI, where the entropy regularized finetuning performs slightly better than the base adaptive VAE-GAN model, with small gains in diversity and quality. We present some examples of the generated sentences in Appendix \ref{app:samples}.

\begin{table}[h!]
\centering
\resizebox{\textwidth}{!}{\begin{tabular}{c|c|c|c}
\hline
Metrics & Standard  & Adaptive & OptAGAN \\ 
\hline
BLEU-2 & 0.880 &  0.886 & \textbf{0.887}  \\
BLEU-3 & 0.675 &  0.683 & \textbf{0.685}  \\
BLEU-4 & 0.448 &  0.456 & \textbf{0.458}  \\
\hline \hline
BBLEU-2 & \textbf{0.860} &  0.854 & 0.854  \\
BBLEU-3 & \textbf{0.666} &  0.660 & 0.661  \\
BBLEU-4 & \textbf{0.453} &  0.449 & 0.451  \\
\hline \hline
FID     & 2.598 &  \textbf{2.539} & 2.562
\end{tabular}}
\caption{YELP results of the automatic metrics for the base VAE-GAN models and the entropy regularized OptAGAN.} 
\label{tab:Table YELP results}
\end{table}

\subsection{Analysis on FID space}\label{sec:fid analysis}
We discuss problems with the FID metric by showing an heatmap of the two-dimensional principal component analysis (PCA) representation and the length distribution of the sentences for the SNLI dataset. Previous works \cite{cit:scratchgan} already investigated the dependency of the FID scores on the length distribution, which can overshadow other problems with the generated samples.

\begin{figure*}[h]
  \floatsetup{floatrowsep=quad}
  \begin{floatrow}[2]
    \ffigbox[.49\textwidth]
  {\caption{PCA representation of the FID space of the SNLI test set (top left), OptAGAN (top right), ScratchGAN (bottom left) and MLE (bottom right) generated data.}\label{fig:snli_fid}}
  {\includegraphics[width=\linewidth]{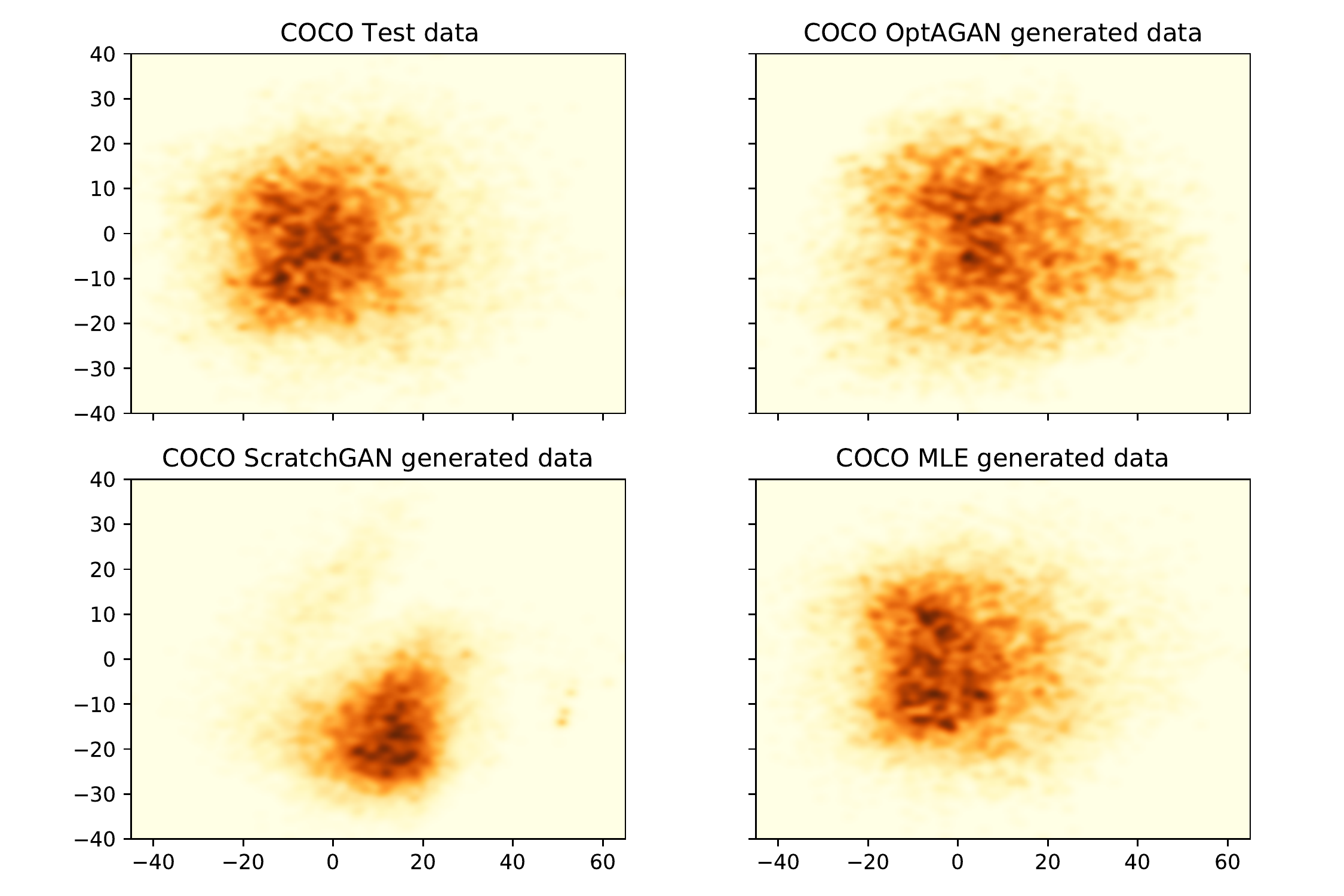}}
    \ffigbox[.49\textwidth]
  {\caption{Length distribution of the four datasets and the FID scores for the SNLI dataset.}\label{fig:snli_length}}
  {\includegraphics[width=\linewidth]{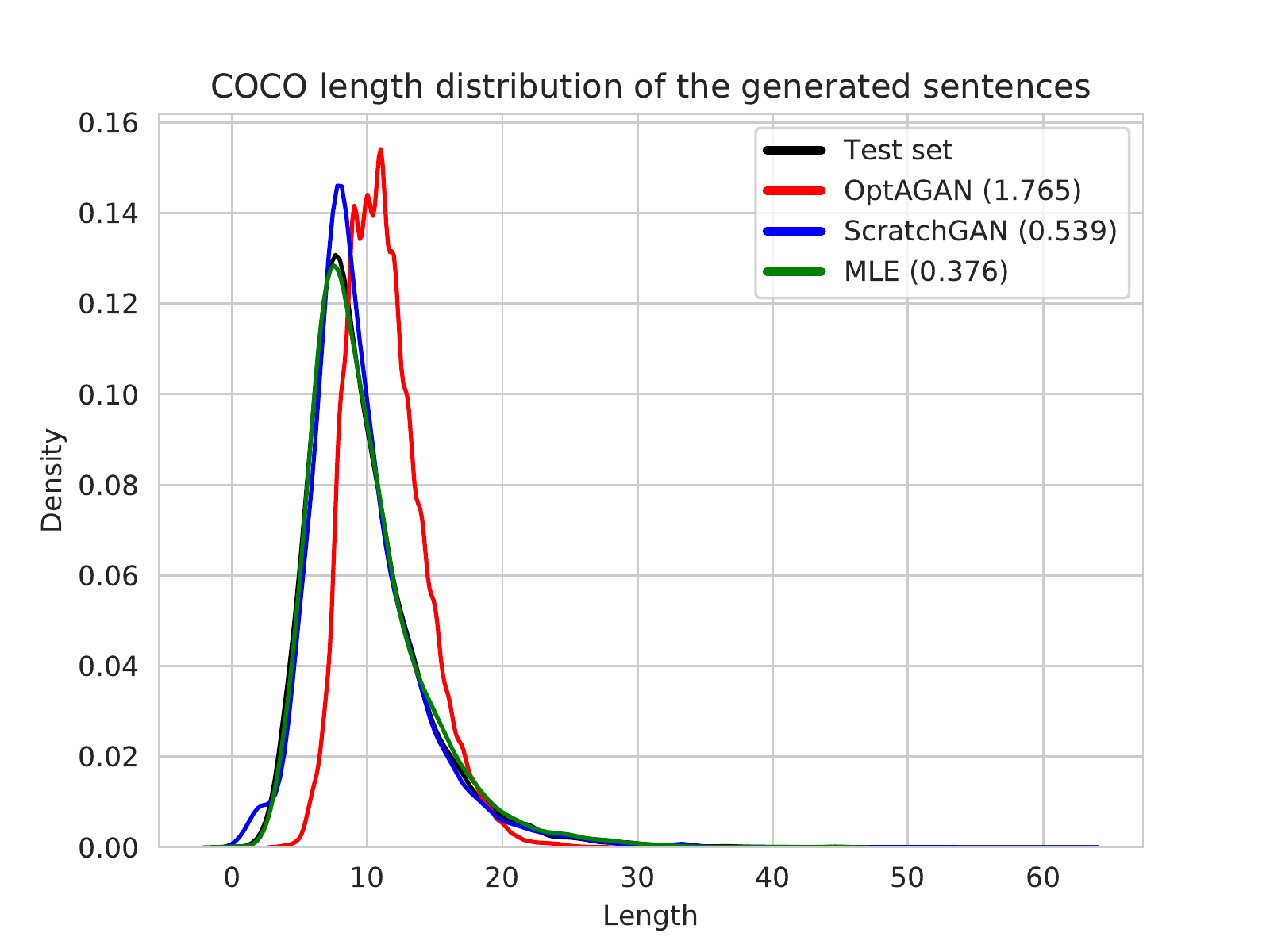}}
  \end{floatrow}
\end{figure*}

We further reinforce those analyses, as a prime example of this issue can be seen in Figure \ref{fig:snli_fid} and \ref{fig:snli_length}, where the FID scores get progressively higher the more the length distribution of the generated sentences is close to the one of the test data, while ignoring the fact that the actual distribution may not match the test one. In fact, the score for the ScratchGAN model is much lower than the one for OptAGAN, although the distribution of the sentences of ScratchGAN in the space misses most of the distribution of the test sentences, as can be seen from the PCA representation. Although only 10-15\% of the overall variability, over the InferSent embedding dimensions, is explained by PCA, there is a huge mismatch that is not addressed by the use of the Fréchet distance, which favours homogeneous length distributions over correct representation of the space. 

\subsection{RL finetuning experiment}\label{sec:rlfinetuning}
In order to fully gauge the strenghts and weaknesses of our entropy penalty approach we set up an experiment where we train a GAN model using the pre-trained Optimus encoder and decoder that are not finetuned on the dataset, so we can finetune a lower quality model. We compare this model with two curiosity-driven regularized model for 1000 and 5000 epochs, respectively. We also use a higher learning rate as we are not interested in preserving the structure of the original sentences and present the results in Table \ref{tab:experiment}. 

\begin{table}[h!]
\centering
\resizebox{\textwidth}{!}{\begin{tabular}{c|c|c|c}
\hline
Metrics & VAEGAN  & RL 1000 & RL 5000 \\ 
\hline
BLEU-2 & 0.641 &  0.707 & \textbf{0.829}  \\
BLEU-3 & 0.370 &  0.444 & \textbf{0.624}  \\
BLEU-4 & 0.198 &  0.243 & \textbf{0.402}  \\
\hline \hline
BBLEU-2 & \textbf{0.752} &  \textbf{0.752} & 0.717  \\
BBLEU-3 & 0.478 &  \textbf{0.497} & 0.479  \\
BBLEU-4 & 0.266 &  \textbf{0.286} & 0.283  \\
\hline \hline
FID     & 2.441 &  \textbf{2.369} &  2.892
\end{tabular}}
\caption{Results of an experiment where we start with an unoptimized VAE-GAN model and finetune it with our RL strategy for 1000 and 5000 epochs on the COCO dataset.}
\label{tab:experiment}
\end{table}

The starting model achieves much worse results than the fully optimized OptAGAN, especially with regards to the BLEU score. After 1000 epochs of RL finetuning, the model improves over both quality and diversity. However, when finetuning for longer, the model cannot balance anymore between exploration and exploitation.
We believe this behaviour is because of the bias that the model has in finding high reward tokens. The RL finetuning does not evaluate all the tokens in the vocabulary, so there might be multiple good scoring tokens which are never considered during the finetuning. This means that our approach is limited by the quality of the original model. 
A countermeasure that might prevent this kind of behaviour from happening could be to increase the penalty for models with higher average entropy, and tuning its value depending on the use case, to further encourage heterogeneous generation.

\section{Conclusions and discussion}
In this work, we benchmark the combination of Optimus and GANs for a text VAE-GAN model, with results that already surpass current methods for the quality of generated texts. We further improve this baseline using entropy-based curiosity-driven rewards to improve both the quality and the diversity of the model. This novel approach could benefit many models utilizing RL for text generation, and supplementary research could be done into exploring advantage policy gradient, or proximal policy optimization with intrinsic rewards. This specification of the reward also allows researchers to prioritize quality, diversity, or to balance between both.
Due to our limited computational resources, we utilize smaller batch sizes than we would otherwise have preferred, as larger batch sizes could help in reducing the high variance gradients of these approaches. Moreover, further research on the automatic metrics could be beneficial not only for evaluation, but also for better reward signals to improve the speed and quality of the finetuning.

\newpage
\appendix
\section{Training details} \label{app:training}
For the finetuning of Optimus, we follow the original work and train for one epoch with the hyperparameters that give the best reconstruction quality, namely:
\begin{itemize}
\setlength\itemsep{-0.2em}
    \item Pre-trained model epoch = 508523
    \item Training epochs\footnote{Due to the size of the COCO dataset, it is the only one where we finetune for 5 epochs.} = 1
    \item Learning rate = $5 \cdot 10^{-5}$
    \item Batch size = 5
    \item Latent size = 768
    \item $\beta$ = 0
    \item Annealing ratio = 0.5
    \item Ratio increase = 0.25
\end{itemize}

We follow a similar approach for the training of the GAN part of the model, where we use standard hyperparameters for the training of WGAN-GP. The best performing epoch of the GAN according to the sum of the BLEU and BBLEU partial scores with 500 texts for both reference and hypothesis is saved.
\begin{itemize}
\setlength\itemsep{-0.2em}
    \item Training epochs = 50
    \item Learning rate = $10^{-4}$
    \item Batch size = 256
    \item Latent size = 768
    \item Maximum sequence length = 100
    \item Number of blocks of generator and discriminator = 10
    \item Gradient penalty $\lambda$ = 10
\end{itemize}	
	
Finally, these are the details for the entropy regularized finetuning. Empirical results showed next to no difference for the BLEU n-gram choice, so we choose 1-gram due to slightly faster computational times, and it also translates into a clearer understanding of the intermediate values. Moreover, we use a small learning rate, in order to keep the same structure as the original sentences.
\begin{itemize}
\setlength\itemsep{-0.2em}
    \item BLEU reward n-grams = 1
    \item Finetuning epochs = 1000
    \item Learning rate = $10^{-6}$ 
    \item Batch size = 32
    \item Epochs value head pre-training = 200
    \item Learning rate value head pre-training = $10^{-4}$ 
\end{itemize}

\section{Adaptive strategy details} \label{app:adap}
\begin{figure}[h!]
\centering
\includegraphics[width=\linewidth]{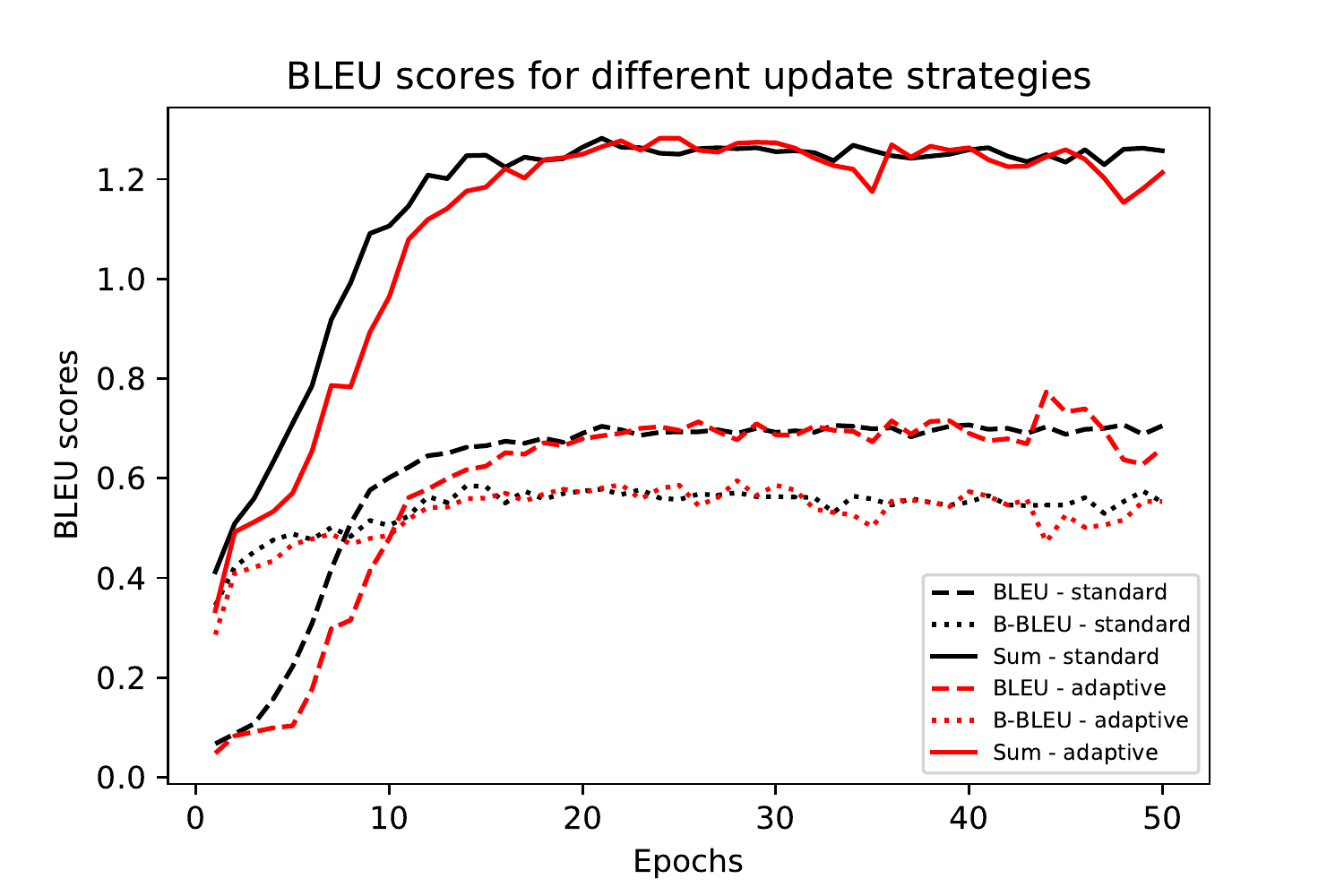}
\caption{Differences in BLEU scores during training for the standard and adaptive update in GANs, evaluated on the COCO validation set.}
\label{fig:norm_adap_coco}
\end{figure}

\begin{figure}[h!]
\centering
\includegraphics[width=\linewidth]{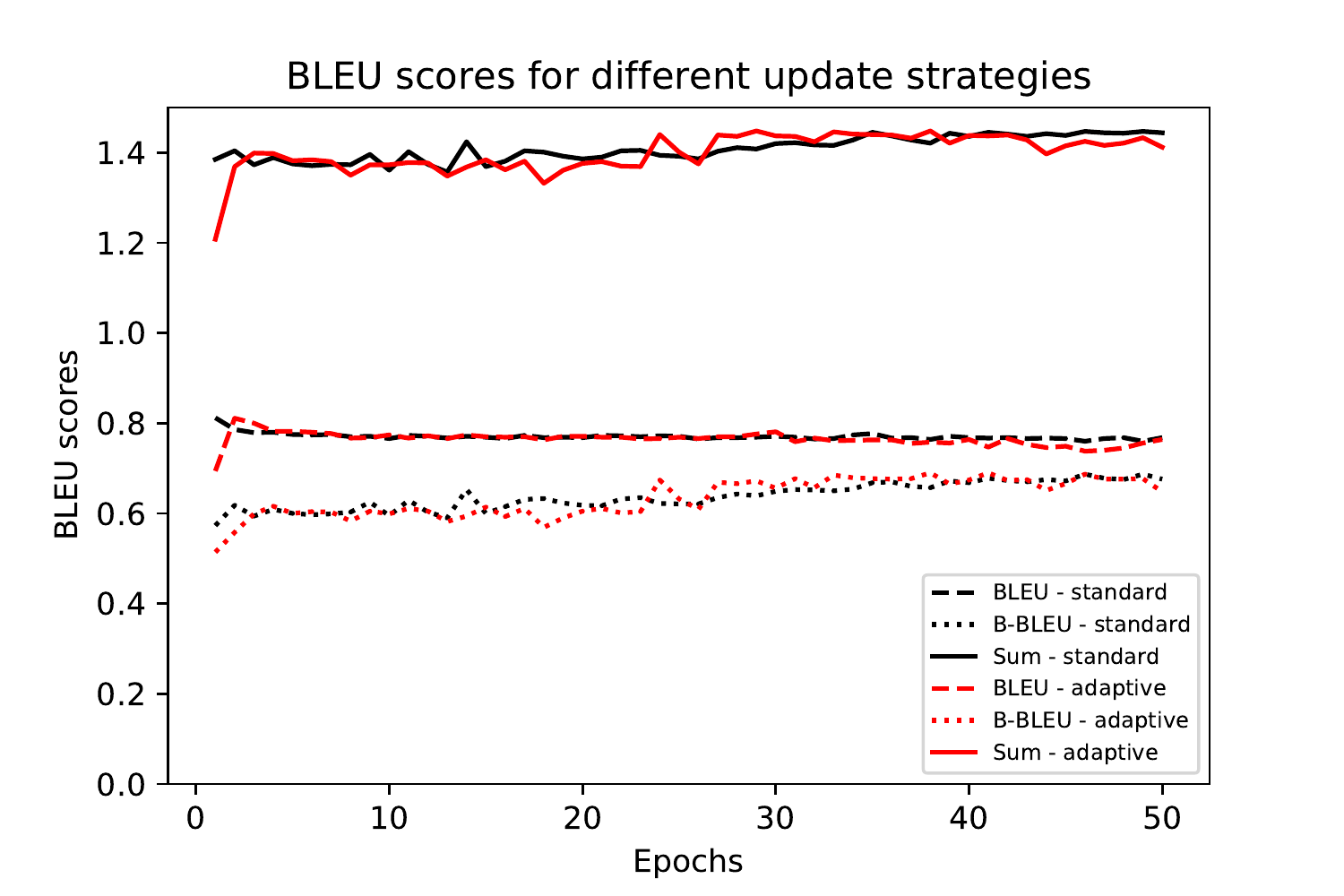}
\caption{Differences in BLEU scores during training for the standard and adaptive update in GANs, evaluated on the YELP validation set.}
\label{fig:norm_adap_yelp}
\end{figure}

Figure \ref{fig:norm_adap_coco} and \ref{fig:norm_adap_yelp} show the validation BLEU and BBLEU for the COCO and YELP datasets. We show the results over these two datasets due to the stark difference between them. 

It is evident that the adaptive strategy is slightly slower to converge. Since the YELP dataset is 10 times larger than COCO, in both cases after about 200,000 samples the two models reach the same quality. The remainder of the training appears to be very stable for the standard update, with little to no changes in both the scores. Meanwhile, the adaptive updates look slightly more volatile, with changes that impact the two scores both positively and negatively, usually following the quality-diversity trade-off principle. 

\section{Generated samples} \label{app:samples}
We include several randomly sampled sentences from OptAGAN for each dataset:
\begin{table*}[h!]
    \centering
    \resizebox{\textwidth}{!}{\begin{tabular}{c|p{0.8\linewidth}|}
    Dataset & Sentences  \\
    \hline
    & a man posing with a bike inside of a forest .  \\
    COCO & a man sitting on a swinging chair pulled by some purple ducks .  \\
    & a woman holding a bird over some flowers on the beach .  \\
    \hline
    & the man is sweating because they are blue .  \\
    SNLI & the man is getting more thank with the dog . \\
    & a man in white clothes stands next to a marketplace where he can store plastic .  \\
    \hline
    & Republican presidential nominee , Donald Trump , has said that 20 to 30 years might be the way to try and narrow it out .  \\
    EMNLP & Whether or not agenda minutes can deliver , it would , therefore , encourage the majority of Scottish MPs to think about that .  \\
    &  Earlier in the day , Bell travelled to Sydney ' s Supreme Court and was effectively blocking a vote of no - one that would produce the album . \\
    \end{tabular}}
    \vspace*{2mm}
    \caption{Examples of the generated unconditional sentences from OptAGAN trained on COCO, SNLI and EMNLP dataset.} 
    \label{tab:Table examples unconditional}
    \end{table*}       

\begin{table*}[h!]
    \centering
    \resizebox{0.99\textwidth}{!}{\begin{tabular}{c|p{0.8\linewidth}|}
    Stars & Generated reviews  \\
    \hline
    1 & horrible experience at this mcdonalds . i have no idea what they are trying to sell you , but if you want anything other than vanilla bean puree and no rice at all , do yourself a favor and go here . for the most part , the waiters do not give a shit .  \\
    2 & i have been to some great buffets , but this was mediocre at best . my husband ordered a turkey sandwich and it was just like anyone else's sandwich . for the service , there wasn't much seating . the food and the \_UNK were stale , awful bread . something to do if you go to vegas for dinner and want to have some classic awesomeness ... maybe try a dim sum instead .  \\
    3 & this place is pretty good . i like the burgers , the selection is pretty good and they have a ginormous amount of steak . being a vegetarian , i took one bite of everything i ordered and went back again . for the friday afternoon rush - they brought me the french fries instead of the turkey , a mousse and mgr .  \\
    4 & loved this place . i had the corned beef burrito , which was very good , though a bit greasy and lacking . they have a good selection of veggie options and happy hour specials and are very attentive to your meal . it's clean , spacious , and the ambiance is great . i have definitely come here when my friends visit vegas to see if they have any other option . the best part about going here is the sitting area outside where you can hang out while eating all you could eat .  \\
    5 & love this place . i have visited all of the great restaurants that offer all sorts of flavors , and to top it all off , all the facilities are extremely clean . \_UNK is the guy . he came for a quick check up , took my dad to the bar and came out free of charge !! seriously ! \\
    \end{tabular}}
    \vspace*{2mm}
    \caption{Examples of the generated sentences from OptAGAN trained on the YELP dataset conditioned on the label.} 
    \label{tab:Table examples YELP}
\end{table*}

\end{document}